\documentclass[10pt,twocolumn,letterpaper]{article}

\usepackage{cvpr}              

\usepackage{multirow}
\usepackage{pgfplots}
\usepgfplotslibrary{groupplots}
\pgfplotsset{compat=1.18}

\DeclareMathOperator*{\argmin}{arg\,min}

\definecolor{cvprblue}{rgb}{0.21,0.49,0.74}
\usepackage[pagebackref,breaklinks,colorlinks,citecolor=cvprblue]{hyperref}

\def\paperID{3976} 
\def\confName{CVPR}
\def\confYear{2024}

\title{KP-RED: Exploiting Semantic Keypoints for Joint 3D Shape \\ Retrieval and Deformation}

\author{Ruida Zhang$^{1*}$, Chenyangguang Zhang$^{1*}$, Yan Di$^{2}$, Fabian Manhardt$^{3}$, \\ Xingyu Liu$^{1}$, Federico Tombari$^{2,3}$, Xiangyang Ji$^{1}$ \\
\textsuperscript{1}Tsinghua University, \textsuperscript{2}Technical University of Munich, \textsuperscript{3}Google\\
{\tt\small \{zhangrd23@mails, zcyg22@mails, xyji@\}.tsinghua.edu.cn}
\thanks{Authors with equal contributions.}
}
\begin{document}
\maketitle

\begin{abstract}
In this paper, we present \textbf{KP-RED}, a unified \textbf{K}ey\textbf{P}oint-driven \textbf{RE}trieval and \textbf{D}eformation framework that takes object scans as input and jointly retrieves and deforms the most geometrically similar CAD models from a pre-processed database to tightly match the target.
Unlike existing dense matching based methods that typically struggle with noisy partial scans, we propose to leverage category-consistent sparse keypoints to naturally handle both full and partial object scans. 
Specifically, we first employ a lightweight retrieval module to establish a keypoint-based embedding space, measuring the similarity among objects by dynamically aggregating deformation-aware local-global features around extracted keypoints.
Objects that are close in the embedding space are considered similar in geometry.
Then we introduce the neural cage-based deformation module that estimates the influence vector of each keypoint upon cage vertices inside its local support region to control the deformation of the retrieved shape.
Extensive experiments on the synthetic dataset PartNet and the real-world dataset Scan2CAD demonstrate that KP-RED surpasses existing state-of-the-art approaches by a large margin.
Codes and trained models are released on \url{https://github.com/lolrudy/KP-RED}.


\end{abstract}

\section{Introduction}

\begin{figure}[t]
\centering
\includegraphics[width=0.49\textwidth]{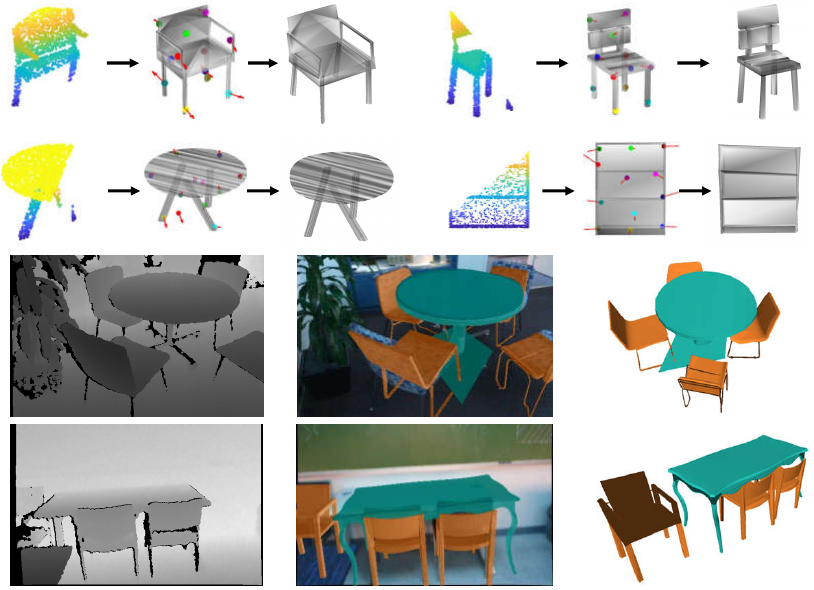}
\caption{\textbf{Top Two Rows:} Given the target point cloud, KP-RED first retrieves the most similar CAD model from the preprocessed database and deforms it to match the target using the keypoints for guidance. \textbf{Bottom Two Rows:} Given a scene scan, KP-RED reconstructs the CAD models of all objects and represents the scene by gathering the reconstructed models.}
\label{fig:teaser}
\end{figure}

Creating high-quality 3D models from noisy object scans has attracted wide research interest \cite{yifan2020neural, jiang2020shapeflow, tatarchenko2019single, kurenkov2018deformnet, wang20193dn} due to its potential applications in 3D scene perception~\cite{nie2020total3dunderstanding, zhang2021holistic}, robotics~\cite{zhai2023sg} and artistic creation~\cite{uy2020deformation, di2023ccd}. 
Previous prior-free methods~\cite{tatarchenko2019single, nie2020total3dunderstanding} directly utilize deep neural networks to recover the object model.
However, due to heavy (self-)occlusion and non-negligible noise, it is often infeasible to infer fine-grained geometric structures without prior knowledge.
To address this issue, \textbf{Retrieval} and \textbf{Deformation} (\textbf{R$\&$D}) methods~\cite{nan2012search,schulz2017retrieval,dahnert2019joint,wang20193dn,uy2021joint,  uy2020deformation,yifan2020neural,jiang2020shapeflow,jakab2021keypointdeformer} are proposed.
These methods first retrieve the most geometrically similar source shape from a certain shape database and then deform the retrieved shape to tightly match the target, yielding a CAD model with fine-grained structural details inherited from the source shape.

However, existing \textbf{R$\&$D} methods typically suffer from two challenges, making them vulnerable to noise and occluded observations.
First, when constructing the embedding space for retrieval, most methods~\cite{uy2021joint, jiang2020shapeflow, gumeli2022roca, li2015joint, dahnert2019joint, di2023ured} resort to single global feature of the input point cloud, which is usually obtained via pooling of point-wise features.
Unfortunately, this strategy inevitably causes the loss of local geometric information, leading to less accurate retrieval, and further deteriorates the deformation quality.
Furthermore, such global feature based embedding is sensitive to occlusion, making it infeasible to handle partial input.
Second, most methods~\cite{jiang2020shapeflow, uy2021joint, wang20193dn, di2023ured} directly utilize dense point matching to control the shape deformation.
However, 
several random outliers in observations may significantly mislead the matching process, resulting in undesired deformation results.

To tackle the aforementioned challenges, we propose KP-RED, a novel keypoint-driven joint \textbf{R$\&$D} framework, which takes a full or partial object scan as input, and outputs the recovered corresponding CAD model via querying a pre-constructed database.
Instead of directly leveraging dense point matching for \textbf{R$\&$D}, we propose to utilize sparse keypoints as the intermediate representation, enabling our unified keypoint-based \textbf{R$\&$D} framework.
Due to lack of ground truth annotations of keypoints, we follow~\cite{jakab2021keypointdeformer} to automatically detect the keypoints in an unsupervised manner by adopting semantically consistent control of shape deformation.
As a result, the discovered keypoints are also proven to be semantically consistent, even under large shape variations across each category.

\begin{figure*}[t]
    \centering
    \includegraphics[width=0.99\textwidth]{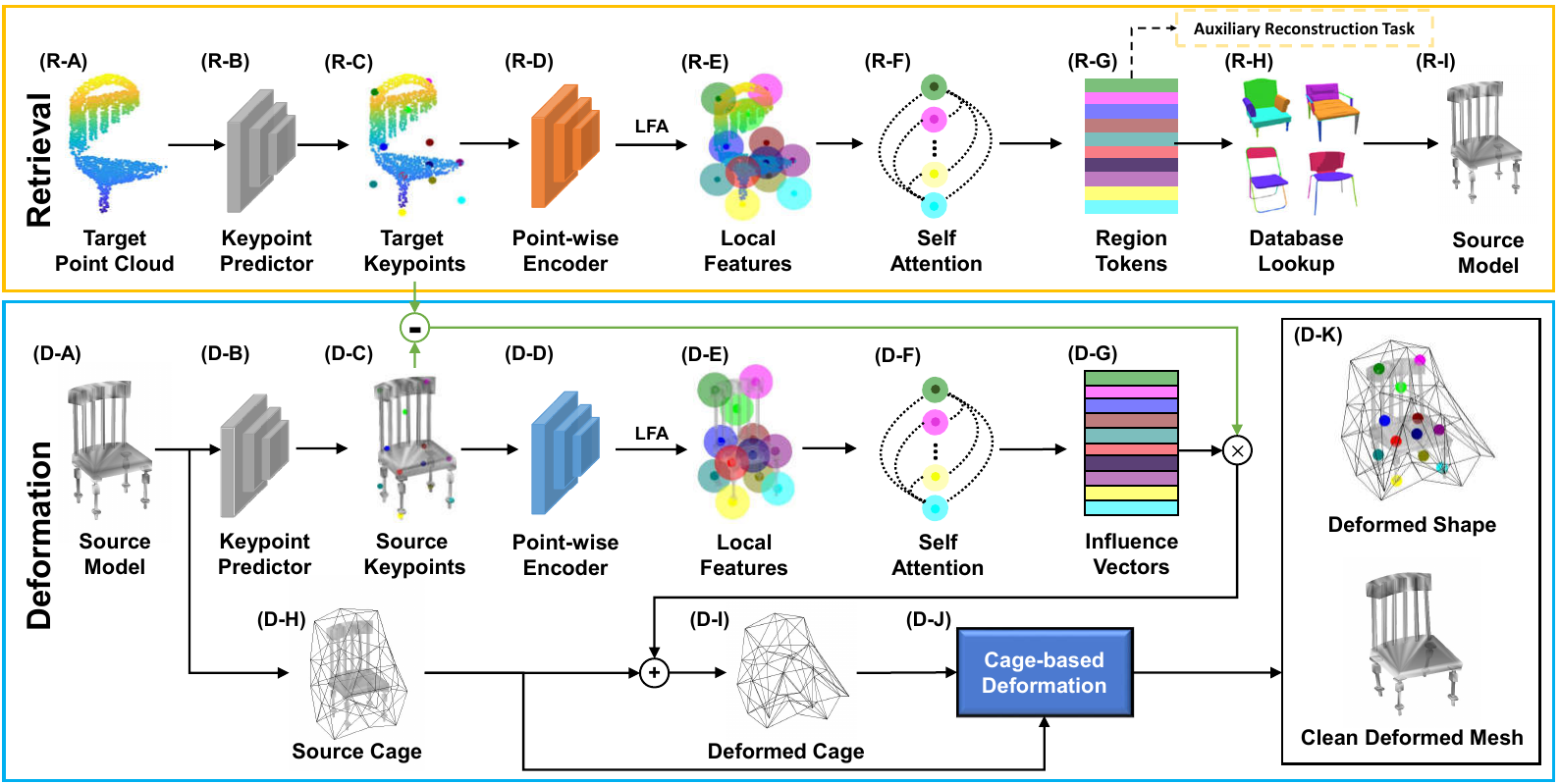}
    \caption{\textbf{Overview of KP-RED.}
    The target point cloud (R-A) is first canonicalized using the estimated pose obtained from an arbitrary pose estimator \cite{di2022gpv,zhang2022ssp,zhang2022rbp,di2021so}, following which the keypoint predictor (R-B) is employed to forecast the target keypoints (R-C).
    An encoder (R-D) predicts point-wise features and Local Feature Aggregation (LFA) is used to obtain the features of each keypoint region (R-E).
    The self-attention module (R-F) extracts the local retrieval token of each region (R-G), which is then compared with the tokens of the database models (R-H).
    The region tokens are supervised with an auxiliary reconstruction task during training.
    The most similar shape to the target is chosen as the source model (R-I).
    The source keypoints are then predicted by the shared keypoint predictor and the local features are extracted via LFA (D-A - D-E). 
    The self-attention module (D-F) predicts the influence vectors (D-G) which demonstrate how the displacements of keypoints inflect the cage.
    Given the cage of the source shape (D-H), the deformed cage (D-I) is derived from the influence vectors.
    Finally, the deformed point cloud and mesh (D-K) are finally computed by the cage-based deformation (D-J).
    }
    \label{fig:pipeline}
\end{figure*}

Specifically, KP-RED consists of two main modules: keypoint-based deformation-aware retrieval and keypoint-driven neural cage deformation.
In the retrieval module (Fig.~\ref{fig:pipeline} Retrieval Block), keypoints are first detected via the keypoint predictor and point-wise features are extracted with PointNet~\cite{qi2017pointnet}.
For each keypoint, we aggregate the point-wise features in its support region (Fig.~\ref{fig:pipeline} (R-E)) to obtain the local retrieval tokens.
Since the locations of keypoints are semantically consistent across each category, we concatenate all local tokens of each object in a uniform order to generate the global retrieval token, which is utilized to retrieve the most similar objects from the database.
In the deformation module (Fig.~\ref{fig:pipeline} Deformation Block), unlike~\cite{jakab2021keypointdeformer, yifan2020neural} that employ global cage scaffolding, we propose to leverage self-attention to simultaneously encapsulate the local fine-grained details and global geometric cues among keypoints so to predict the influence vector upon the support region of each keypoint, which is then interpolated onto the source shape to control the deformation.

Compared with dense matching based baselines~\cite{uy2021joint, jiang2020shapeflow, di2023ured}, our keypoint-based framework holds two main advantages.  
First, the extracted keypoints are semantically consistent across each category, allowing effective occlusion reasoning and noise suppression.
Thereby KP-RED can better handle noisy partial object scans than competitors.
Second, our keypoint-based feature aggregation approach preserves fine-grained local geometry information, yielding a more accurate embedding space for retrieval.

Our main contributions are summarized as follows,
\begin{itemize}
\setlength{\itemsep}{0pt}
\setlength{\parsep}{0pt}
\setlength{\parskip}{1pt}
    \item 
    We present a unified network KP-RED for 3D shape generation from object scans, which learns category-consistent keypoints to jointly retrieve the most similar source shape from the pre-established database and control the shape deformation.
    \item
    We design a keypoint-driven local-global feature aggregation scheme to establish the shape embedding space for retrieval, which performs effectively for both full and partial object scans, enabling multiple real-world applications.
    \item
    We introduce a novel cage-based deformation scheme with self-attention that uses keypoints to control the local deformation of the retrieved shape.
\end{itemize}

\section{Related Works}

\textbf{Neural Shape Generation.}
Recent advances in neural networks have led to the development of generative latent representations for 3D shapes. 
\cite{mescheder2019occupancy,chen2019learning,park2019deepsdf, remelli2020meshsdf,jang2021codenerf,mildenhall2021nerf,xu2022point,zhai2024commonscenes,ddfho} model geometry as implicit functions, while \cite{achlioptas2018learning,sun2020pointgrow,yang2019pointflow,mo2019structurenet, xie2019pix2vox, wang2018pixel2mesh} generate point clouds, voxels or meshes to model shapes explicitly. 
Factorized representations are studied by \cite{li2017grass,gao2019sdm}, decomposing shapes into different geometric parts and handling of the geometric variations of each part separately.
However, while these methods exhibit impressive representation abilities, they may struggle to preserve structural details and to handle complex objects due to the lack of prior knowledge. 

\textbf{CAD Model Retrieval.}
Retrieving a CAD model that closely matches a 3D scan of a real-world object is a critical issue in 3D scene understanding. 
While many prior works directly retrieve the most similar CAD model by evaluating similarity in the descriptor space \cite{schulz2017retrieval, bosche2008automated} or the latent embedding space of neural networks \cite{li2015joint,dahnert2019joint, gumeli2022roca, avetisyan2019end}, the direct retrieval may not always yield satisfying results since the model database cannot contain all instances. 
To address this limitation, recent works propose extracting deformation-aware embeddings \cite{uy2020deformation} or developing novel optimization targets \cite{ishimtsev2020cad} to better fit the details of the target shape after deformation. 
However, their deformation modules are fixed and non-trainable, leading to inferior performance.

\textbf{3D Shape Deformation.}
One of the fundamental issues of geometry processing is to deform a source 3D model to tightly match a target shape.
Traditional methods~\cite{sorkine2007rigid,ganapathi2018parsing,huang2008non} directly optimize the deformed shapes to fit the targets.
However, they are only applicable to complete target shapes and do not generalize well to the real-world scenarios since the object scans are typically partial due to (self-)occlusion.
By modeling deformation as volumetric warps~\cite{kurenkov2018deformnet, jack2018learning}, cage deformations~\cite{yifan2020neural,jakab2021keypointdeformer}, vertex-based offsets~\cite{wang20193dn}, or flows~\cite{jiang2020shapeflow}, recent approaches attempt to learn deformation priors from a set of shapes by neural networks.
Cage-based deformation~\cite{yifan2020neural} is particularly noteworthy for its ability to preserve geometry details, and \cite{jakab2021keypointdeformer} extends it by adopting automatically discovered semantic keypoints to enable human users to control the shape explicitly. 
However, most of these works do not study the retrieval process.
Only a few works~\cite{uy2021joint,jiang2020shapeflow} jointly study \textbf{R$\&$D}.
Uy et al.~\cite{uy2021joint} designs a novel training strategy to jointly optimize retrieval and deformation modules, but its deformation module depends on the part annotations of the database models which are labor intensive to obtain.
U-RED \cite{di2023ured} proposes point-wise residual-guided retrieval metrics and a one-to-many module to handle noisy and partial inputs.
However, it also depends on the part annotations.
ShapeFlow~\cite{jiang2020shapeflow} constructs a flow-based deformation space and utilizes an auto-decoder to extract features for retrieval.
Nevertheless, the auto-decoder needs to perform time-consuming online optimization during inference.
Thus, we propose a keypoint-based joint \textbf{R$\&$D} framework KP-RED which yields high-quality CAD models with no requirements of extra annotations and runs in real-time.

\section{KP-RED}
In this section, we first present the overview of KP-RED, and then introduce our \textbf{R$\&$D} method for processing full shapes in detail in Sec.~\ref{sec:deform}, Sec.~\ref{sec:retrieval}.
In Sec.~\ref{sec:partial}, we demonstrate our confidence-based dynamic feature aggregation technique for partial shape.
The \textbf{Overview} of KP-RED is shown in Fig.~\ref{fig:pipeline}.
Given an input full or partial object scan, KP-RED first constructs a keypoint-guided deformation-aware embedding space to retrieve the most similar model from the database, and then deforms the model to match the input shape via keypoint-driven cage deformation.

In the \textbf{Retrieval} module, given the target point cloud $S_{tgt}$ as input, the keypoint detector (R-B) predicts $N_K$ semantic keypoints $\mathbf{K}_{tgt} = \{K^{(1)}_{tgt}, K^{(2)}_{tgt}, ..., K^{(N_K)}_{tgt}\}$ on $S_{tgt}$.
Then PointNet~\cite{qi2017pointnet} is employed to extract point-wise features.
For each keypoint $K^{(i)}_{tgt}$, we aggregate its corresponding deformation-aware local feature ${l}^{(i)}_{tgt}$ by pooling the point features within a ball region $\mathcal{R}^{(i)}_{tgt}$ centered at $K^{(i)}_{tgt}$, where $i=1,...,N_K$ (R-E).
Self-attention is employed to discover the region-to-region relations and predict the local retrieval tokens of each keypoint region $\{\mathcal{T}^{(1)}_{tgt}, \mathcal{T}^{(2)}_{tgt}, ..., \mathcal{T}^{(N_K)}_{tgt}\}$ (R-G).
Since the locations of keypoints are semantically consistent across each category, we concatenate all local tokens 
in a uniform order to generate the global deformation-aware token $\mathcal{T}_{tgt}$.
We then utilize $\mathcal{T}_{tgt}$ to compare with the tokens in the database, so to retrieve the most similar shape as the source shape (R-I).

In the \textbf{Deformation} module, the source shape $S_{src}$ (D-A) is fed into the identical keypoint detector to extract keypoints $\mathbf{K}_{src} = \{K^{(1)}_{src}, K^{(2)}_{src}, ..., K^{(N_K)}_{src}\}$ from $S_{src}$. 
Influence vectors $\{I_1, I_2, ..., I_{N_K}\}$ (D-G) are predicted via the self-attention module (D-F), describing how each keypoint influences its support cage vertices (Fig.~\ref{fig:deform}~(a)).
Finally, we obtain the deformed source shape $S_{src2tgt}$ from the deformed cage (D-I) by adopting the mean value coordinate based interpolation approach~\cite{ju2005mean}.

\subsection{Keypoint-Driven Deformation \label{sec:deform}}
In the deformation module, we aim to deform the retrieved source shape $S_{src} \in \mathbb{R}^{N_P \times 3}$ to tightly match the input target shape $S_{tgt}\in \mathbb{R}^{N_P \times 3}$.
Note that $N_P$ denotes the number of points.
We follow~\cite{yifan2020neural, ju2005mean, jakab2021keypointdeformer} and adopt the keypoint-driven neural-cage based deformation and extend it with self-attention to encapsulate local structural details and global geometric cues.


\noindent \textbf{Neural Cage Deformation.}
To control the deformation of the source shape $S_{src}$, we adopt the sparse cage scaffolding strategy~\cite{yifan2020neural, jakab2021keypointdeformer}.
First, a coarse control mesh (cage) with vertices $C_{src} \in \mathbb{R}^{N_{C} \times 3}$ is computed to enclose $S_{src}$, so that displacements of $C_{src}$ can be interpolated to any point on $S_{src}$, via constructing mean value coordinates~\cite{yifan2020neural}, enabling to control the deformation of $S_{src}$.
Second, 
we predict the influence vector $I_i \in \mathbb{R}^{N_{C} \times 1}$ of each keypoint $K^{(i)}_{src} \in \mathbb{R}^{1 \times 3}$ in $\mathbf{K}_{src}$ (Fig.~\ref{fig:pipeline} (D-G), Fig.~\ref{fig:deform} (a2)).
This vector describes how the cage vertices are influenced by the displacements of the keypoints.
Finally, the differences between the keypoints $\mathbf{K}_{tgt}$ of the target shape $S_{tgt}$ and $\mathbf{K}_{src}$ of $S_{src}$ are compared to guide the algorithm moving $C_{src}$.
The resulting deformed cage vertices $C_{src2tgt}$ are calculated as
\begin{equation}
    C_{src2tgt} = C_{src} + \sum_{i=1}^{N_K} I_i (K_{tgt}^{(i)} - K_{src}^{(i)}).
\label{eq:ctgt}
\end{equation}
By interpolating $C_{src2tgt}$ on $S_{src}$~\cite{yifan2020neural, jakab2021keypointdeformer}, we obtain $S_{src2tgt}$ that tightly matches $S_{tgt}$.

\noindent \textbf{Geometric Self-Attention.}
Previous methods~\cite{yifan2020neural,jakab2021keypointdeformer} directly utilize the global feature to predict the influence vectors, resulting in unsatisfactory performance due to inevitable loss of local information.
We instead adopt local feature aggregation and self-attention mechanism to capture local and global information and to, thus, preserve more geometric details.
To this end, we first predict the point-wise features via PointNet~\cite{qi2017pointnet} and gather the local features $l^{(i)}_{src}$ of each keypoint $K^{(i)}_{src}$ by pooling the point-wise features inside its support ball region centered at $K^{(i)}_{src}$ with radius $r$ (Fig.~\ref{fig:pipeline} (D-E), Fig.~\ref{fig:deform} (a1)).
We then use a self-attention module to discover region-to-region relations and inject global information to $l^{(i)}_{src}$, whilst preserving the local information.
Finally, the influence vector of each keypoint is derived from the feature of its own support region.
We restrict the influence vector to only influence 
cage vertices inside the local support region of the keypoint.
This ensures that the local information around the keypoint is fully exploited, leading to finer-grained deformation.
Moreover, the region-to-region relations complement essential global information to the local features.
For example, when facing a partial scan of a symmetric object (\textit{e.g.} chair), the structure of the missing part can be inferred from its corresponding symmetric regions.
Thus, the geometric self-attention not only preserves structural details, yet also enhances robustness towards partial inputs.

\noindent \textbf{Training.} 
During training, our final objective is composed of two loss terms, used to simultaneously supervise the learning of keypoint extraction and shape deformation.
The former term is supposed to enforce shape similarity $\mathcal{L}_{sim}$ by calculating the Chamfer Distance between $S_{src2tgt}$ and $S_{tgt}$. 
The other term is meant to regularize the keypoints~\cite{jakab2021keypointdeformer}. 
This terms encourages keypoints to be well-distributed by minimizing the Chamfer Distance between different keypoints and $N_K$ points sampled by means of Farthest Point Sampling on $S_{src}$.
The overall loss function of the deformation module is thus defined as
\begin{equation}
    \mathcal{L}_{def} = \mathcal{L}_{sim} + \lambda_{kpt} \mathcal{L}_{kpt},
\label{eq:ldef}
\end{equation}
where $\lambda_{kpt}$ weights the contribution of the keypoints regularization term. 
More details on the definitions of loss terms are provided in the Supplementary Material.

\subsection{Deformation-Aware Retrieval \label{sec:retrieval}}

\begin{figure}[t]
    \centering
    \includegraphics[width=0.45\textwidth]{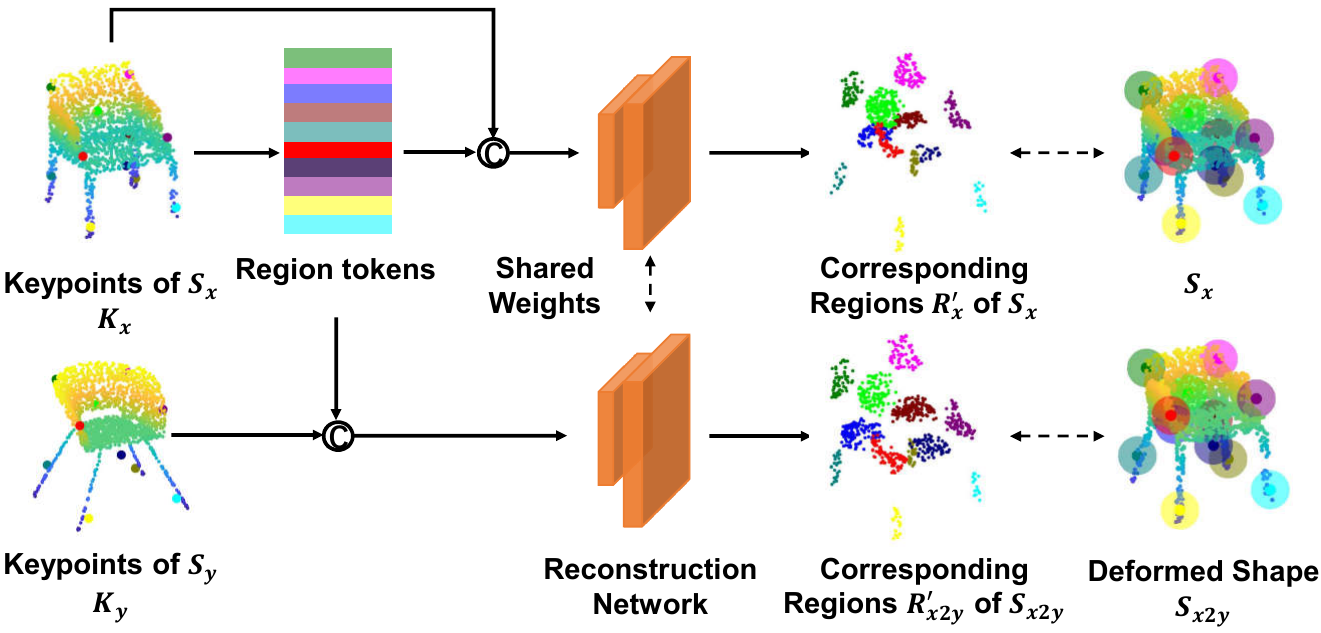}
    \caption{The training procedure of the retrieval module. Given the keypoints $\mathbf{K}_x$ and the region tokens extracted from the shape $S_x$, the reconstruction network reconstructs the corresponding regions $R'_x$ of $S_x$.
    Meanwhile, the network reconstructs the regions of the deformed shape $R'_{x2y}$ from the region tokens and $\mathbf{K}_y$.
    }
    \label{fig:retrieval}
        \vspace{-0.3cm}

\end{figure}

The retrieval module aims to retrieve the most geometrically similar source shape $S_{src}$ for the input target shape $S_{tgt}$ from a pre-constructed database.
The retrieval task faces two main challenges.
First, the retrieval process should be deformation-aware, meaning the retrieved shape should match the target shape tightly after deformation.
Second, the overall retrieval module should be lightweight and real-time with minimal additional computational cost.
To address these challenges, we design a novel keypoint-based retrieval method.


\noindent \textbf{Local-Global Feature Embedding.}
Unlike~\cite{uy2021joint} that leverages a directly learned global feature for retrieval, we instead utilize local-global keypoint-based features.
As shown in Fig.~\ref{fig:pipeline}, given the target object $S_{tgt}$, we predict its keypoints $\mathbf{K}_{tgt}$ by the keypoint detector.
Similar to the deformation module, we adopt local feature aggregation and self-attention to extract
the local features $\mathcal{T}^{(i)}_{tgt}$ of each keypoint $K^{(i)}_{tgt}$ as the local retrieval tokens.
Since the keypoints are, as aformentioned, semantically consistent across each category, we concatenate all local tokens $\{\mathcal{T}^{(1)}_{tgt}, \mathcal{T}^{(2)}_{tgt}, ..., \mathcal{T}^{(N_K)}_{tgt}\}$ in a uniform order to generate the global deformation-aware token $\mathcal{T}_{tgt}$.
During inference, we choose the source model as 
\begin{equation}
    S_{src} = \argmin_{\omega \in  \Omega} f_{\mathcal{L}_1}(\mathcal{T}_{tgt}, \mathcal{T}_{\omega}),
\end{equation}
where $\Omega$ denotes the pre-established model database, $T_{\omega}$ is the global token of the database model $\omega$ and $f_{\mathcal{L}_1}(\cdot,\cdot)$ computes the $\mathcal{L}_1$ distance between the two tokens.

\noindent \textbf{Training.}
We illustrate the full training procedure in Fig.~\ref{fig:retrieval}.
To supervise the learning of retrieval tokens $\mathcal{T}_{tgt}$ and  $\mathcal{T}_{src}$, we introduce a novel auxiliary reconstruction task.
Given two randomly selected shapes $S_x$ and $S_y$ from the training set, we first extract their keypoints $\mathbf{K}_{x}$, $\mathbf{K}_{y}$ and retrieval tokens $\mathcal{T}_{x}$, $\mathcal{T}_{y}$, and then utilize the deformation module to deform $S_x$ to match $S_y$, yielding $S_{x2y}$.
Subsequently, we adopt an MLP-based reconstruction network $\Psi$, which processes two tasks in parallel, $\Psi_1: (\mathcal{T}^{(i)}_{x}, \mathbf{K}_{x}) \rightarrow {R_{x}^{(i)}}'$ and $\Psi_2:(\mathcal{T}^{(i)}_{x}, \mathbf{K}_{y}) \rightarrow {R_{x2y}^{(i)}}'$, where ${R_{x}^{(i)}}'$ and ${R_{x2y}^{(i)}}'$ denote the reconstruction results of the support region  of the \textit{i}th keypoint of $S_x$ and $S_{x2y}$ (${R_{x}^{(i)}}$ and ${R_{x2y}^{(i)}}$) respectively.
Therefore, our training objective is defined as
\begin{equation}
    \mathcal{L}_{ret} = \frac{1}{N_K} \sum_{i=1}^{N_K} (f_{CD}({R_{x}^{(i)}}, {R_{x}^{(i)}}') + f_{CD}({R_{x2y}^{(i)}},{R_{x2y}^{(i)}}')),
    \label{lret}
\end{equation}
with $f_{CD}$ denoting the Chamfer Distance between two point clouds.
The first reconstruction task $\Psi_1$ forces $\mathcal{T}_{x}$ to encapsulate the geometric information of $S_x$.
On the other hand, comparing $\Psi_2$ with the deformation module that prescribes $(S_{x}, \mathbf{K}_{y}) \rightarrow S_{x2y}$ via neural cage deformation, where $S_{x}=\cup_{i=1}^{N_K} R_{x}^{(i)}$ and $S_{x2y}=\cup_{i=1}^{N_K} R_{x2y}^{(i)}$. Thereby, $\mathcal{T}_{x}$ is encouraged to also capture cues related to the deformation of $S_x$.
We divide the reconstruction task into different regions to force each token to capture the local structural details of each support region and enhances  the granularity of retrieval results.

\subsection{Handling Partial Point Cloud \label{sec:partial}}
\begin{figure}[t]
    \centering
    \includegraphics[width=0.46\textwidth]{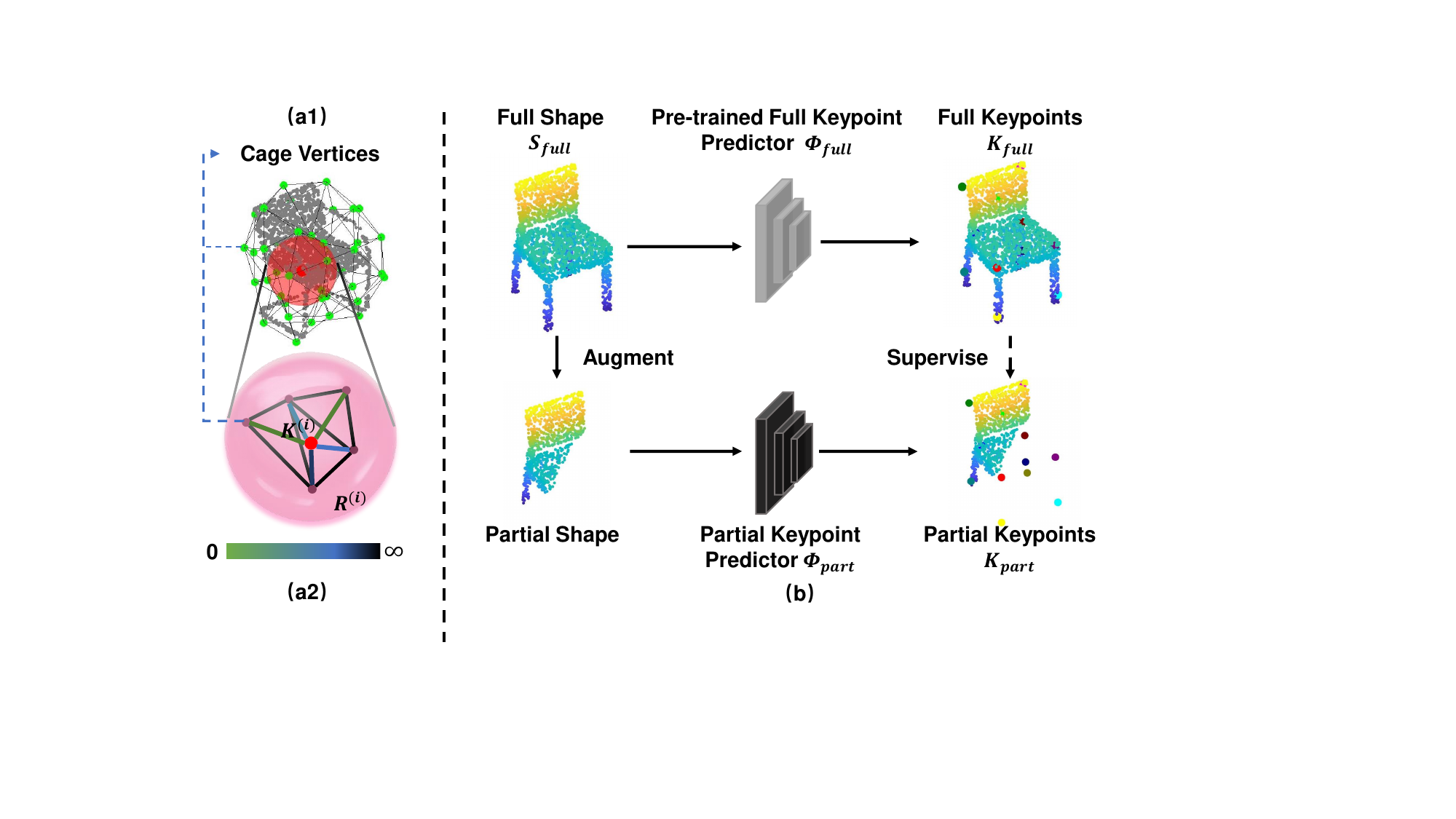}
        \vspace{-0.2cm}
    \caption{\textbf{(a1)}: The support region $R^{(i)}$ of the specific keypoint $K^{(i)}$.
    \textbf{(a2)}: The influence vectors $I_i$ of the specific keypoint  $K^{(i)}$.
    The color indicates the influence weight of the keypoint towards each cage vertex (as in Eq.~\ref{eq:ctgt}).
    \textbf{(b)}: The training procedure of the keypoint predictor for partial shapes. We employ the keypoint predictor trained with full shapes for supervision.
    }
    \vspace{-0.3cm}
    \label{fig:deform}
\end{figure}

Due to (self-)occlusion, poor lighting conditions, viewpoints, \textit{etc}, actual real-world scans are oftentimes just partially available. 
In order to simulate real-world partial condition, we thus augment full shapes in PartNet for generation of partial shapes by means of random slicing.
Please refer to the Supplementary Material for details.

Compared with full shape \textbf{R$\&$D}, handling partial shapes poses a new challenge. In particular, the observed point cloud is typically non-uniformly distributed in 3D space, rendering it difficult to reliably extract keypoints from poorly observed regions.
To address this limitation, we thus propose a confidence-based dynamic feature extraction, to improve robustness towards partial inputs.

Without additional priors, we assume that the point density is a good measurement to understand the reliability of point observations in most cases.
Specifically, if the density in the support region $R^{(i)}$ of keypoint $K^{(i)}$ is low, then $K^{(i)}$ can be considered unreliable and should contribute less for \textbf{R$\&$D}.
Conversely, $K^{(i)}$ should be assigned a larger weight.
We define the density $D_i$ at each keypoint $K^{(i)}$ as the normalized average density in its support region $R^{(i)}$, $D_i=\min(N^{(i)}_R/(\alpha V), 1)$, where $N^{(i)}_R$ denotes the number of points in $R^{(i)}$, $V$ describes the region volume, and $\alpha$ is a constant for normalization.
The density of all keypoints is $\mathbf{D} = \{D_1, D_2, ..., D_{N_K}\} \in \mathbb{R}^{N_K}$.

\noindent \textbf{Retrieval.}
We use the aforementioned density $\mathbf{D}$ as the confidence weight to select the source model $S_{src}$ from the database $\Omega$ via 
\begin{equation}
    S_{src} = \argmin_{\omega \in \Omega} \sum_{i=1}^{N_K} D_i f_{\mathcal{L}_1}(\mathcal{T}_{tgt}^{(i)}, \mathcal{T}_{\omega}^{(i)}).
\end{equation}
Thereby, keypoint regions with higher density contribute more to the final retrieval results.

\noindent \textbf{Deformation.}
As shown in Fig.~\ref{fig:deform} (b), we introduce an additional keypoint predictor $\Phi_{part}$ to handle partial shapes. 
We use the keypoint predictor $\Phi_{full}$ (corresponding to (R-B) in Fig.~\ref{fig:pipeline}) trained with full shapes to guide the learning of $\Phi_{part}$ in a teacher-student manner, where only the parameters in $\Phi_{part}$ are updated. 
In essence, given an augmented partial shape $S_{part}$ and its corresponding full shape $S_{full}$, we obtain the keypoints of the partial shape $\mathbf{K}_{part}=\Phi_{part}(S_{part})$ and the full shape $\mathbf{K}_{full}=\Phi_{full}(S_{full})$.
As $S_{part}$ is augmented from $S_{full}$, they are required to also possess identical keypoints.
Moreover, keypoints with higher confidence should contribute more to the deformation result and thus require stronger supervision signals.
We use again the density $\mathbf{D}$ as the confidence weight and define the weighted keypoint loss as
\begin{equation}
    \mathcal{L}_{wkpt} = \sum_{i=1}^{N_K} D_i f_{\mathcal{L}_1}({K}_{full}^{(i)}, {K}_{part}^{(i)}).
\end{equation}

Since the common Chamfer Distance (CD) as defined in Eq.~\ref{eq:ldef} is bilateral, it is not a suitable metric to evaluate shape similarity for partial shapes. Therefore, we replace it with the Unilateral Chamfer Distance (UCD) from the target shape towards the deformed shape for the similarity loss, denoted as $\mathcal{L}_{usim}$.
We refer again to the Supplemental Material for additional details.



Finally, the overall loss function of the deformation module, for handling partial shapes, is defined as
\begin{equation}
    \mathcal{L}_{pdef} = \mathcal{L}_{usim} + \lambda_{wkpt} \mathcal{L}_{wkpt},
\end{equation}
where $\lambda_{wkpt}$ is used as the weighting parameter.

\section{Experiments}


\textbf{Datasets.}
To evaluate the effectiveness of our method, we leverage a synthetic datasets PartNet \cite{mo2019partnet}
and a real-world dataset Scan2CAD~\cite{avetisyan2019scan2cad}.
For PartNet, we adopt the same split of database, training set and test set as in~\cite{uy2021joint}.
The shapes in PartNet come from ShapeNet \cite{chang2015shapenet}.
PartNet contains 1419 models in database, 11433 instances for training and 2861 for testing.
Please note that we do not need the part annotations like in \cite{uy2021joint} and only use the mesh models for training.
Scan2CAD~\cite{avetisyan2019scan2cad} is a real-world dataset developed upon ScanNet~\cite{dai2017scannet} and provides the ground truth masks, poses and corresponding CAD models for 14225 objects.
The input point clouds in Scan2CAD are generated by back-projecting the depth maps.
We first centralize the point clouds and then follow \cite{uy2021joint} to canonicalize them by the provided rotations.
We conduct experiments on the \{chair, table, cabinet\} categories on PartNet and Scan2CAD.
To further evaluate the reconstruction quality when facing partial inputs under different occlusion ratios, we augment shapes in PartNet to generate partial shapes with random slicing (see Supp. Mat.).
We generate partial target point clouds with occlusion ratio of $25\%$, $50\%$ and $75\%$ from the test split of Partnet as the test set.

\textbf{Implementation Details.}
Following \cite{uy2021joint}, $N_P=2048$ points are sampled from each shape and normalized into a unit cube to serve as the input of the network.
We first train the deformation module from scratch and then utilize the predicted keypoints to train the retrieval module.
The source and target shapes are randomly selected from the model database and the training set in the above mentioned two steps.
To handle the partial shape, we train the partial keypoint predictor while freezing other parameters learned from full shapes.
During training, the target shape is augmented by random slicing with the occlusion ratio $\gamma$ uniformly sampled from $\gamma \sim \mathcal{U}(25\%, 90\%)$.
The baseline models \cite{uy2021joint, jiang2020shapeflow} trained with full shapes are fine-tuned with the same augmentation for partial shape evaluation.
The original Chamfer Distance loss is also replaced by the Unilateral Chamfer Distance (UCD) loss for fair comparison.
We directly utilize the models (KP-RED and \cite{uy2021joint, jiang2020shapeflow, di2023ured}) trained on PartNet dataset to inference on the validation set of Scan2CAD, and the model database remains the same as the setting of PartNet.
We use $N_K=12$ keypoints for all categories.
The radius of the support region is set to $r=0.3$. 
The parameters for all loss terms are selected empirically and kept unchanged in experiments unless specified, with $\{\lambda_{kpt}, \lambda_{wkpt}\}=\{2, 20\}$.
We run all experiments on a single NVIDIA 3090 GPU and employ the Adam optimizer~\cite{kingma2014adam} with batch size of 16 and base learning rate of 1e-3 for deformation module and 1e-2 for retrieval module.
We train the two modules for 30 epochs each, about 300K iterations.
Detailed descriptions of network architectures are provided in the Supplemental Material.

\textbf{Evaluation Metrics.}
The typical Chamfer Distance (CD) between the reconstructed model and the object scanning is used for full shape evaluation, while the Unilateral Chamfer Distance (UCD) is reported for partial shapes.
On Scan2CAD dataset, we use the ground truth model to compute the UCD metric. 
All methods retrieve the top 10 source candidates and choose the best \textbf{R$\&$D} result to calculate metrics as in U-RED~\cite{di2023ured}.
We also provide results using top 1, 5, 25 retrieval candidates in Sup. Mat..
The \textit{average} metrics in all tables are obtained via averaging over all \textbf{instances}.

\begin{figure}[t]
    \centering
    \includegraphics[width=0.44\textwidth]{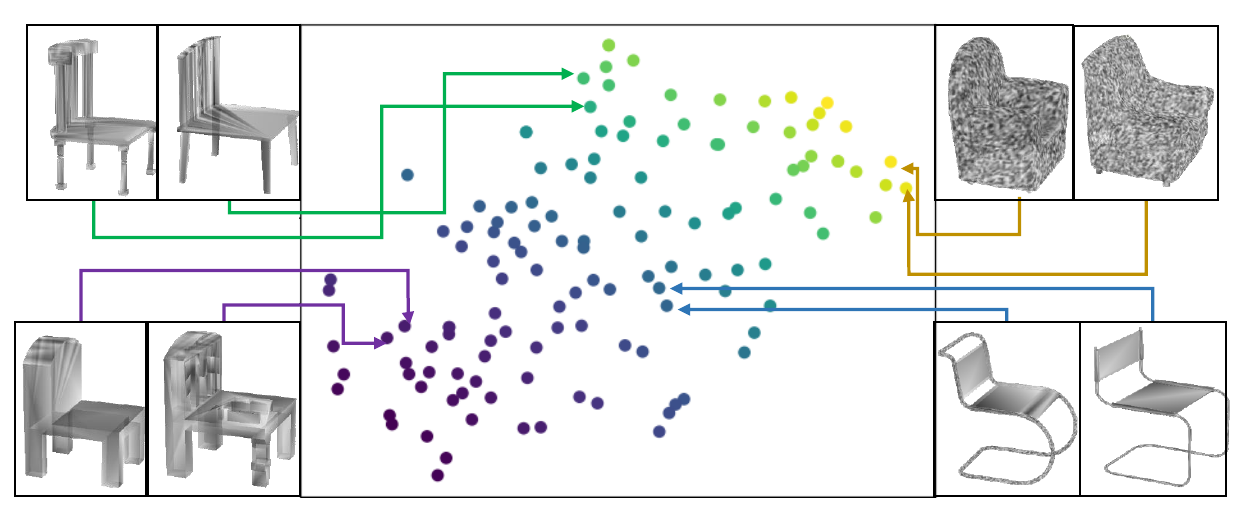} 
        \vspace{-0.2cm}
    \caption{The visualization of the learned retrieval tokens of database shapes via t-SNE \cite{van2008visualizing}. 
    Objects whose tokens are close in the embedding space are considered similar in geometry.}
    \label{fig:tsne}
    \vspace{-0.3cm}
\end{figure}

\subsection{Experiments on Full Shapes}

\begin{table}[t]
\begin{center}
\begin{tabular}{c|ccc|c}
\hline
Method            & Chair          & Table         & Cabinet &Average \\
\hline
ShapeFlow \cite{jiang2020shapeflow} &0.238   &0.400   &0.514   &0.340
 \\
 Uy \textit{et al.} \cite{uy2021joint} & 0.182 & 0.170 & 0.179 & 0.166
 \\
 U-RED \cite{di2023ured} & 0.238 & 0.088 & 0.123 & 0.143 \\
Ours        & \textbf{0.091} & \textbf{0.084} & \textbf{0.109} & \textbf{0.089}\\
\hline
\end{tabular}
\end{center}
\vspace{-0.5cm}
\caption{Chamfer Distance metrics for joint \textbf{R$\&$D} results on full shapes on PartNet dataset \cite{mo2019partnet}. 
Overall best results are \textbf{in bold}.
}
\label{tab:full_result}
\end{table}

\begin{figure}[t]
    \centering
    \includegraphics[width=0.48\textwidth]{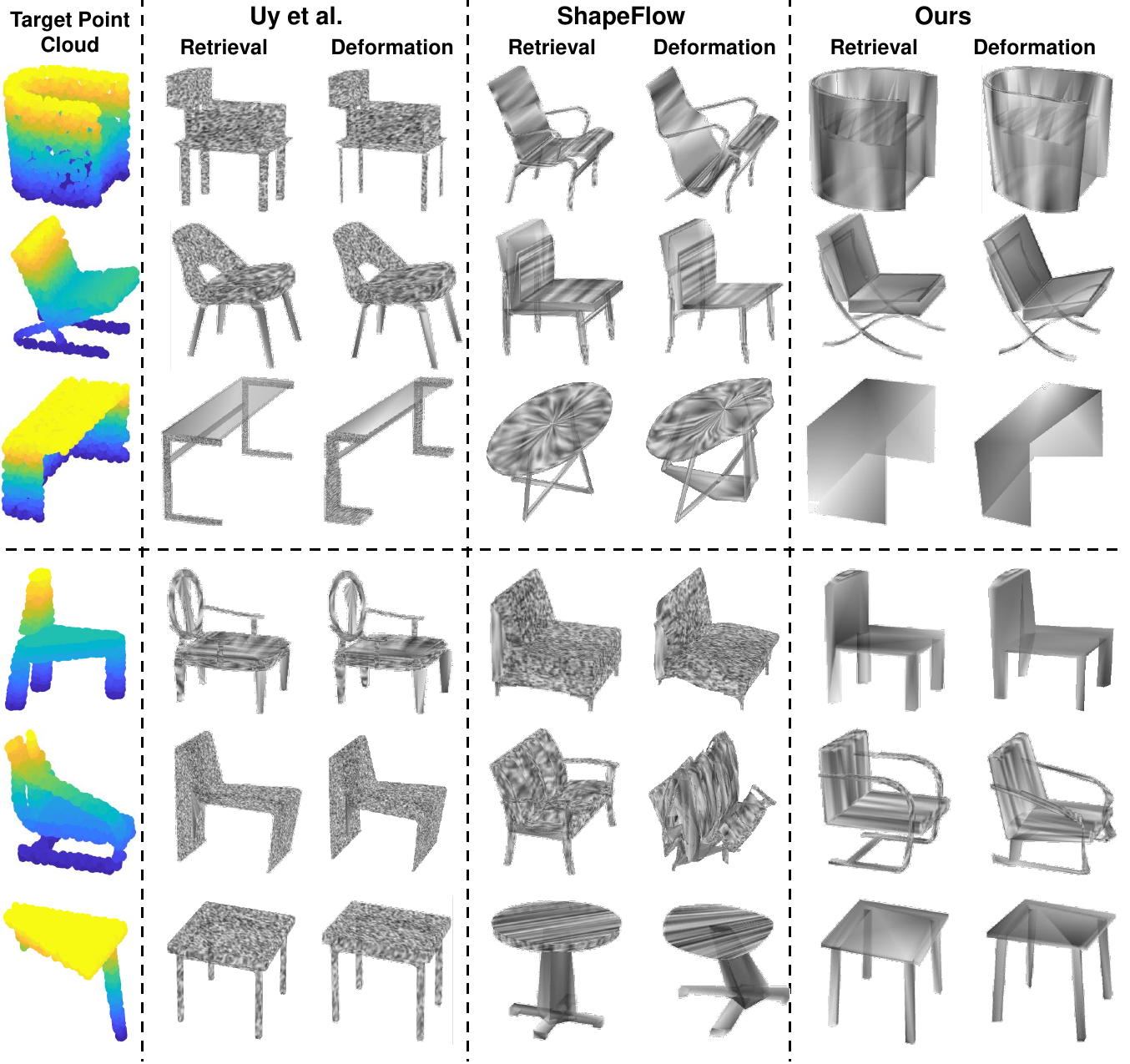}
    \caption{Qualitative \textbf{R$\&$D} results on PartNet~\cite{mo2019partnet}.
    \textbf{Top Block:} Full shape \textbf{R$\&$D}. 
     \textbf{Bottom Block:} Partial shape \textbf{R$\&$D}.}
    \label{fig:full_vis}
        \vspace{-0.3cm}
\end{figure}

To demonstrate the joint \textbf{R$\&$D} ability of KP-RED, we compare our method with the state-of-the-art \cite{uy2021joint, jiang2020shapeflow}, and present the results in Tab. \ref{tab:full_result}.
KP-RED consistently outperforms other competitors in all categories and datasets under the Chamfer Distance metrics, demonstrating our superior ability of \textbf{R$\&$D} under different conditions.
In particular, on PartNet dataset, we obtain superior results with a relative improvement of 73.8\%  under the \textit{average Chamfer Distance}, compared with ShapeFlow~\cite{jiang2020shapeflow}.
When comparing with the second best method Uy \etal \cite{uy2021joint}, our improvement is still significant with an relative improvement of 37.8\%. 

Moreover, the inference time of Uy \textit{et al.} \cite{uy2021joint} reaches 0.7 seconds per instance, while ShapeFlow~\cite{jiang2020shapeflow} adopts online optimization for better performance, which leads to non-negligible computational expenses and very long inference time, about 45 seconds per instance.
In contrast, KP-RED maintains a real-time inference speed with about 30 ms per instance, whilst surpassing both by a large margin in terms of \textbf{R$\&$D} quality. 
Fig.~\ref{fig:tsne} demonstrates that our unsupervised keypoint-driven retrieval module is capable of successfully establishing an embedding space for measuring the similarity among objects.
As illustrated in Fig.~\ref{fig:full_vis}, our \textbf{R$\&$D} results are clearly more similar to the target compared to other methods.
We attribute this to our well-designed retrieval module, 
employing local-global feature embedding for effective comparison among objects.
Moreover, the cage-based deformation with geometric self-attention preserves structural details and boosts deformation quality.
In Supplementary Material, we perform an oracle retrieval experiment to illustrate the effectiveness of our keypoint-guided deformation module.



\subsection{Experiments on Partial Shapes}

\definecolor{darkyellow}{RGB}{232, 155, 0}

\begin{figure}

\begin{tikzpicture}
\begin{groupplot}[
    group style={
      {group size=2 by 2, vertical sep=24pt}},
    height=0.2\textwidth,width=0.25\textwidth,y label style={below=1mm}, x label style={below=-2mm}, title style={below=-2mm},
    ymin=0, ymax=0.5,
    ytick={0, 0.1, 0.2, 0.3, 0.4}
    ]

    \nextgroupplot[title=Average,  
    legend style={at={(1.07,-2.18)},
      anchor=south,legend columns=-1, },
    ] 

\addplot[
    color=darkyellow,
    mark=x,
    ]
    coordinates {
    (0, 0.183)
    (25, 0.210)
    (50, 0.314)
    (75, 0.335)
    };
    \addlegendentry{SF~\cite{jiang2020shapeflow}}
    
    \addplot[
    color=blue,
    mark=o,
    ]
    coordinates {
    (0, 0.083)
    (25, 0.179)
    (50, 0.241)
    (75, 0.270)
    };
    \addlegendentry{Uy~\cite{uy2021joint}}

    \addplot[
    color=green,
    mark=triangle,
    ]
    coordinates {
    (0, 0.074)
    (25, 0.180)
    (50, 0.255)
    (75, 0.306)
    };
    \addlegendentry{U-RED~\cite{di2023ured}}

\addplot[
    color=red,
    mark=square,
    ]
    coordinates {
    (0, 0.046)
    (25, 0.049)
    (50, 0.056)
    (75, 0.066)
    };
    \addlegendentry{Ours}
    
    \nextgroupplot[title=Chair]
\addplot[
    color=darkyellow,
    mark=x,
    ]
    coordinates {
    (0, 0.168)
    (25, 0.186)
    (50, 0.271)
    (75, 0.301)
    };

\addplot[
    color=green,
    mark=triangle,
    ]
    coordinates {
    (0, 0.121)
    (25, 0.203)
    (50, 0.264)
    (75, 0.315)
    };
    
    \addplot[
    color=blue,
    mark=o,
    ]
    coordinates {
    (0, 0.091)
    (25, 0.145)
    (50, 0.194)
    (75, 0.232)
    };

\addplot[
    color=red,
    mark=square,
    ]
    coordinates {
    (0, 0.048)
    (25, 0.050)
    (50, 0.056)
    (75, 0.061)
    };
    
    \nextgroupplot[title=Table, xlabel={Occlussion Ratio [\%]}]
    \addplot[
    color=darkyellow,
    mark=x,
    ]
    coordinates {
    (0, 0.182)
    (25, 0.217)
    (50, 0.352)
    (75, 0.363)
    };

\addplot[
    color=green,
    mark=triangle,
    ]
    coordinates {
    (0, 0.045)
    (25, 0.168)
    (50, 0.227)
    (75, 0.264)
    };
    
    \addplot[
    color=blue,
    mark=o,
    ]
    coordinates {
    (0, 0.085)
    (25, 0.191)
    (50, 0.272)
    (75, 0.292)
    };

\addplot[
    color=red,
    mark=square,
    ]
    coordinates {
    (0, 0.043)
    (25, 0.046)
    (50, 0.056)
    (75, 0.069)
    };
    
    \nextgroupplot[title=Cabinet, xlabel={Occlussion Ratio [\%]},]
    \addplot[
    color=darkyellow,
    mark=x,
    ]
    coordinates {
    (0, 0.283)
    (25, 0.297)
    (50, 0.293)
    (75, 0.337)
    };

\addplot[
    color=green,
    mark=triangle,
    ]
    coordinates {
    (0, 0.067)
    (25, 0.137)
    (50, 0.240)
    (75, 0.349)
    };

    \addplot[
    color=blue,
    mark=o,
    ]
    coordinates {
    (0, 0.092)
    (25, 0.293)
    (50, 0.301)
    (75, 0.344)
    };
    

\addplot[
    color=red,
    mark=square,
    ]
    coordinates {
    (0, 0.060)
    (25, 0.060)
    (50, 0.061)
    (75, 0.070)
    };
  \end{groupplot}

\end{tikzpicture}
\caption{Unilateral Chamfer Distance metrics for joint \textbf{R$\&$D} results on the augmented partial PartNet~\cite{mo2019partnet}. SF stands for ShapeFlow \cite{jiang2020shapeflow}.}
\label{fig:partial_result}
\end{figure}
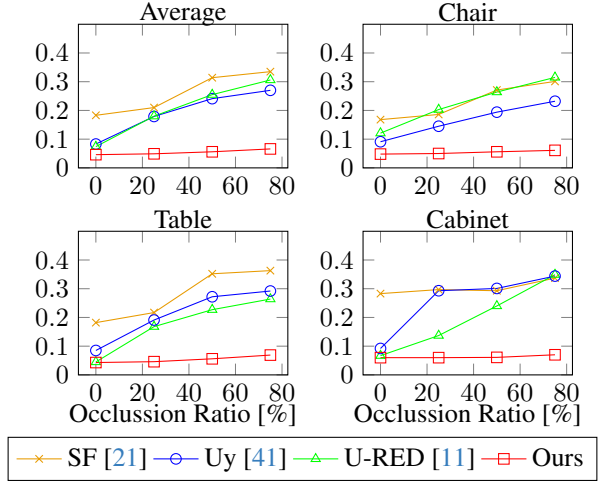

\begin{table}[t]
\begin{center}
\begin{tabular}{c|ccc|c}
\hline
Method            & Chair          & Table         & Cabinet &Average\\
\hline
ShapeFlow \cite{jiang2020shapeflow} &0.230   &0.302   &0.345   & 0.265
 \\
Uy \textit{et al.} \cite{uy2021joint} & 0.158 & 0.190 & 0.676 & 0.210
\\
U-RED \cite{di2023ured} & 0.227   & 0.132   & 0.316   & 0.207
 \\
Ours        &\textbf{0.059}  &\textbf{0.057}  &\textbf{0.073}  &\textbf{0.060} \\
\hline
\end{tabular}
\end{center}
\vspace{-0.5cm}
\caption{Unilateral Chamfer Distance metrics for joint \textbf{R$\&$D} results on Scan2CAD \cite{avetisyan2019scan2cad}. 
Overall best results are \textbf{in bold}.
}
\label{tab:scan2cad_result}
        \vspace{-0.3cm}
\end{table}

\begin{figure}[t]
    \centering
    \includegraphics[width=0.45\textwidth]{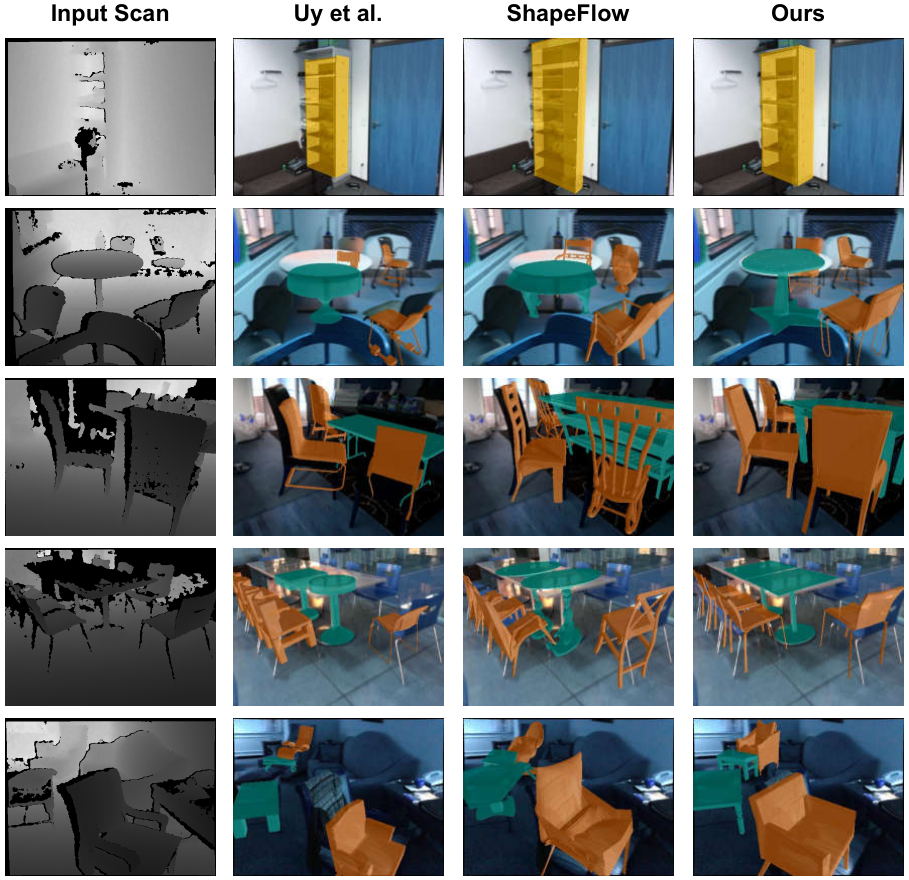}
            \vspace{-0.1cm}
    \caption{Qualitative results on Scan2CAD dataset~\cite{avetisyan2019scan2cad}. The \textbf{R$\&$D} results are rendered on the RGB images for better visualization.}
    \label{fig:vis_scan2cad}
        \vspace{-0.2cm}
\end{figure}

We conduct experiments on two datasets, namely real-world Scan2CAD \cite{avetisyan2019scan2cad} and synthetic augmented PartNet \cite{mo2019partnet}, to comprehensively demonstrate the robustness of KP-RED for handling partial point clouds.
As shown in Tab.~\ref{tab:scan2cad_result} and Fig.~\ref{fig:partial_result}, KP-RED constantly outperforms the current state-of-the-art \cite{uy2021joint, jiang2020shapeflow} in both real-world and synthetic scenarios by a significant margin.
On Scan2CAD dataset, under real-world occlusion, KP-RED exceeds Uy \textit{et al.}~\cite{uy2021joint}, U-RED and ShapeFlow by  71.4\% 71.0\%,  and 77.3$\%$ under the UCD error respectively.
Fig.~\ref{fig:vis_scan2cad} shows superior visualization quality of KP-RED with accurate retrieval and accurate deformation.
On partial PartNet, when the occlusion ratio increases from 0\% to 75\%, the \textit{average} UCD error of KP-RED only increases by 0.020, while the error of U-RED and ShapeFlow increases 0.187 and 0.232, respectively.
As can be seen KP-RED is more robust to incomplete input due to the dynamic feature extraction.
More qualitative results are shown in the Supplementary Material.


\subsection{Ablation Studies}

\begin{table}
\begin{center}
\scalebox{0.9}
{
\begin{tabular}{c|ccccccc}
\hline
& GSA & DAR   & LGF  & Chair & Table & Cabinet & Avg.     \\
\hline
(a)&$\times$ & $\times$ & $\times$ &0.135  &0.187  &0.126  &0.161 \\
(b)&$\times$ & $\checkmark$ & $\times$ &0.128  &0.187  &0.124  &0.158 \\
(c)&$\checkmark$ & $\times$ & $\times$ &0.116  &0.105  &0.117  &0.110 \\
(d)&$\checkmark$ & $\times$ & $\checkmark$ &0.099  &0.088  &0.113  &0.094 \\
(e)&$\checkmark$ & $\checkmark$ & $\times$ &0.103  &0.092  &0.113  &0.098 \\
(f)&$\checkmark$ & $\checkmark$ & $\checkmark$ & \textbf{0.091} & \textbf{0.084} & \textbf{0.109} & \textbf{0.089}\\
\hline
\end{tabular}
}
\end{center}
\vspace{-0.5cm}
\caption{Ablation studies on full shapes of PartNet \cite{mo2019partnet}. Avg. denotes the average CD metric.
GSA denotes Geometric Self-Attention.
Without GSA, we use PointNet~\cite{qi2017pointnet} to extract the global feature of the input shape, like in~\cite{uy2021joint}.
DAR denotes Deformation-Aware Retrieval.
We only use the reconstruction task $\Psi_1$ to train the retrieval network when it is ablated.
LGF denotes Local-Global Feature Embedding for the retrieval process.
When it is ablated, we directly utilize the global feature for retrieval.
}
    \vspace{-0.4cm}

\label{tab:full_ablation}
\end{table}
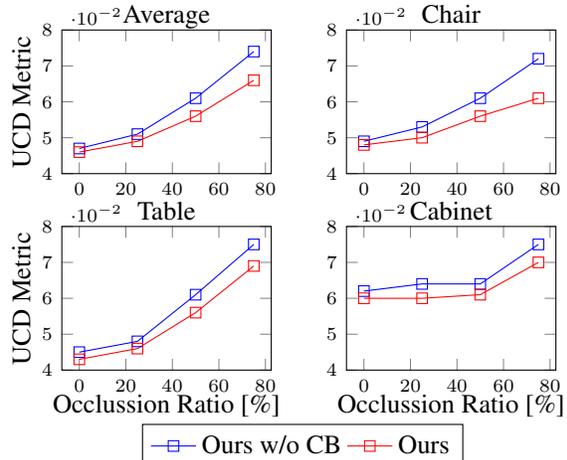
\begin{figure}
\begin{tikzpicture}
\begin{groupplot}[
    group style={
      {group size=2 by 2,
      vertical sep=20pt}},
    height=0.2\textwidth,width=0.25\textwidth,y label style={below=-4mm}, x label style={below=-2mm}, title style={below=-2.5mm},
    ticklabel style={font=\scriptsize},
    ymin=0.04, ymax=0.08
    ]

    \nextgroupplot[title=Average,  
    legend style={at={(1.15,-2.05)},
      anchor=south,legend columns=-1, },
    ylabel={UCD Metric}] 
    \addplot[
    color=blue,
    mark=square,
    ]
    coordinates {
    (0, 0.047)
    (25, 0.051)
    (50, 0.061)
    (75, 0.074)
    };
    \addlegendentry{Ours w/o CB}

\addplot[
    color=red,
    mark=square,
    ]
    coordinates {
    (0, 0.046)
    (25, 0.049)
    (50, 0.056)
    (75, 0.066)
    };
    \addlegendentry{Ours}
    
    \nextgroupplot[title=Chair]
\addplot[
    color=blue,
    mark=square,
    ]
    coordinates {
    (0, 0.049)
    (25, 0.053)
    (50, 0.061)
    (75, 0.072)
    };

\addplot[
    color=red,
    mark=square,
    ]
    coordinates {
    (0, 0.048)
    (25, 0.050)
    (50, 0.056)
    (75, 0.061)
    };
    
    \nextgroupplot[title=Table, xlabel={Occlussion Ratio [\%]},ylabel={UCD Metric}]
    \addplot[
    color=blue,
    mark=square,
    ]
    coordinates {
    (0, 0.045)
    (25, 0.048)
    (50, 0.061)
    (75, 0.075)
    };

\addplot[
    color=red,
    mark=square,
    ]
    coordinates {
    (0, 0.043)
    (25, 0.046)
    (50, 0.056)
    (75, 0.069)
    };
    \nextgroupplot[title=Cabinet, xlabel={Occlussion Ratio [\%]},]
    \addplot[
    color=blue,
    mark=square,
    ]
    coordinates {
    (0, 0.062)
    (25, 0.064)
    (50, 0.064)
    (75, 0.075)
    };

\addplot[
    color=red,
    mark=square,
    ]
    coordinates {
    (0, 0.060)
    (25, 0.060)
    (50, 0.061)
    (75, 0.070)
    };
  \end{groupplot}

\end{tikzpicture}
\vspace{-0.3cm}
\caption{Ablation studies of \textbf{C}onfidence-\textbf{B}ased Dynamic Feature Extraction (CB) on partial PartNet~\cite{mo2019partnet}.}
    \vspace{-0.5cm}

\label{fig:partial_ablation}
\end{figure}

We conduct ablation studies in both full shape (Tab.~\ref{tab:full_ablation}) and partial shape (Fig.~\ref{fig:partial_ablation}) scenarios.
The ablations on full shapes mainly aim at verifying the effectiveness of our proposed \textbf{G}eometric \textbf{S}elf-\textbf{A}ttention (GSA) in Sec.~\ref{sec:deform}, \textbf{D}eformation-\textbf{A}ware \textbf{R}etrieval (DAR) and \textbf{L}ocal-\textbf{G}lobal \textbf{F}eature Embedding (LGF) in Sec.~\ref{sec:retrieval}.
While on partial shapes, \textbf{C}onfidence-\textbf{B}ased Dynamic Feature Extraction (CB)  in Sec.~\ref{sec:partial} is ablated.

Tab.~\ref{tab:full_ablation} exhibits the ablations of GSA, DAR and LGF, conducting on full shapes of PartNet.
Comparing (a) and (c) in Tab.~\ref{tab:full_ablation}, our proposed encoder with GSA contributes to an improved \textit{average} performance by about $32\%$.
The effectiveness of DAR is demonstrated by (c) and (e) with a reduction of \textit{average} CD error by $10\%$.
After incorporating LFA and DAR, utilizing LGF further enhances the \textit{average} performance by $9\%$ ((e) and (f)).
As described in Sec.~\ref{sec:partial}, the effectiveness of \textbf{R$\&$D} with \textbf{C}onfidence-\textbf{B}ased Dynamic Feature Extraction (CB) is illustrated in Fig.~\ref{fig:partial_ablation}.
When ablating CB, we assume that the confidence weights of all keypoints are equal.
There is a general trend that CB contributes more when the occlusion ratio is high.
CB improves the \textit{average} performance by $11\%$ for up to $75\%$ occlusion and $8\%$ for less than $50\%$ occlusion.
This indicates that CB is an essential strategy for handling partial shape.

\section{Conclusion}

In this paper, we introduce KP-RED, a unified framework for 3D shape generation from full or partial object scans. 
Our approach employs category-consistent keypoints to jointly retrieve the most geometrically similar shapes from a pre-constructed database and deform the retrieved shape to tightly match the input. 
We propose a keypoint-driven local-global feature aggregation scheme to extract deformation-aware features for retrieval, and a neural cage-based deformation algorithm to control the local deformation of the retrieved shape. 
In the future, we plan to extend our technique to 3D scene understanding. 

\noindent\textbf{Acknowledgement.}
This work was supported by the National Key R\&D Program of China under Grant 2018AAA0102801.

{\small
\bibliographystyle{ieee_fullname}
\bibliography{egbib}
}

\end{document}